\documentclass[11pt]{article}
\usepackage{cmap}
\usepackage[margin=1in]{geometry}
\usepackage[T1]{fontenc}
\usepackage{lmodern}
\usepackage{natbib}
\usepackage{booktabs}
\usepackage{array}
\usepackage{amsmath,amssymb}
\usepackage{xcolor}
\usepackage{hyperref}
\hypersetup{colorlinks=true,linkcolor=blue,citecolor=blue,urlcolor=blue}
\usepackage{caption}
\usepackage{enumitem}
\usepackage{xurl}

\title{\textbf{Which Answer Did the Seventeen-Year-Old Write?}\\
\large A Behavioral Case Study of Author-Age Misattribution, Miscalibrated Confidence,\\ and Non-Corrective Self-Explanation in Six Frontier Language Models}

\author{Brianne Lee, M.S., M.P.H. \\ \small Independent Researcher \\ \small \texttt{briannelee011@gmail.com}}
\date{October 2026}

\begin{document}
\maketitle

\begin{abstract}
Three students answered the same exam prompt by hand: \emph{``What is happiness? Are you happy?''} Their ages, recorded at collection, are 17, 28, and 45. The 17-year-old's answer is hedged and formally phrased. Asked to match each answer to an age and gender label, most frontier language models confidently assign it to an adult. This paper reports a structured case study of that failure across six models: Claude Fable~5, Gemini~3.1~Pro, GPT-5.6~Sol, Grok~4.5, GPT-4o~(2024), and DeepSeek~V4. We ran 1{,}081 API sessions and scored 3{,}225 analysis-eligible responses across six experimental conditions that manipulate framing, disclosure format, and audience-awareness instruction. Three of six models perform at or near chance ($1/3$) on the matching task while reporting 73--88\% mean confidence. The youngest writer's answer is misattributed to an adult in up to 90\% of trials. Removing the multiple-choice scaffold collapses accuracy further: three models never place within three years of the true age once open-ended estimation is required. Institutional framing (\emph{``a community-college biology final''}) significantly degrades the best-performing model's accuracy; a neutral label does not. A single audience-awareness instruction significantly improves one model (+21 points) and moves another to ceiling, with no detectable effect on the three weakest models. Stated confidence is uncorrelated or inversely correlated with correctness for five of six models. After ground truth is revealed, one model names its own stereotyping in 68 of 114 explanations, yet shows the lowest accuracy and highest confidence of any model under blind presentation: articulate retrospective diagnosis that does not track the quality of its judgments. We report effect sizes, confidence intervals, and significance tests throughout. This is a three-item, single-topic, single-language case study meant to isolate a mechanism, not to estimate a population bias rate. Data, code, and raw model outputs are released under CC~BY~4.0 / MIT.
\end{abstract}

\section{Introduction}

Text carries two kinds of information about its author: what they said, and how they said it. The second kind (register, hedging, sentence complexity, vocabulary) is exactly what stylometric and forensic-linguistic research treats as a reliable behavioral signature \citep{grieve2023register}. This paper argues it is also a signature that large language models routinely mistake for a demographic one. Formality reads as maturity. Hedged uncertainty reads as adult self-awareness. Casual humor reads as youth, independent of who actually produced the text.

The concrete case motivating this study: on a community-college biology final, an extra-credit item asked students to answer, in their own words, \emph{``What is happiness? Are you happy?''} Three students answered by hand. Their ages were recorded at collection: 17 (male), 28 (female), 45 (male). One answer reads, in part, that happiness is a concept with many interpretations depending on the person, and that the writer is content with life but unsure if they are happy. Presented with all three answers and asked which was written by whom, most of the frontier models tested here confidently guess that this answer belongs to an adult in their late twenties or forties. It is the 17-year-old's.

This is a counter-stereotypical stimulus by construction. A teenager writing formally, for a teacher, produces exactly the register that both human folk theory and, as this paper shows, model behavior associate with adulthood. That construction is the point. Studies that infer bias from scraped or synthetic text with self-reported or inferred demographic labels cannot isolate cases where the true label runs directly against the surface cue. Such cases are rare in unfiltered natural data. A small set of real, human-written, counter-stereotypical items, studied exhaustively rather than broadly, can.

\paragraph{Contributions.} This paper makes six contributions:
\begin{enumerate}[nosep]
\item Real, human-written, counter-stereotypical ground truth (three handwritten exam answers with ages recorded at collection) in place of scraped or synthetic text with inferred labels.
\item A six-condition framing manipulation (blind, institutional label, neutral label, audience-awareness instruction, open-ended attribution, and self-reported probability distribution) that isolates \emph{what kind} of context helps or harms attribution accuracy.
\item A direct within-stimulus contrast between forced-choice matching and open-ended attribution, showing that multiple-choice scaffolding substantially masks failure severity.
\item A trait-versus-signal decomposition of model confidence: between-model differences in stated confidence dwarf within-model correctness signal, with formal significance testing per model.
\item A post-hoc self-explanation measure: coded text after ground truth is revealed, compared against the same model's accuracy and confidence in its other, independent sessions.
\item Full transparency about a meta-level finding: AI-assisted tooling used during the study itself produced several confident, fluent errors, each caught by human verification before publication.
\end{enumerate}

We report this work as a structured case study. With three stimulus items on one topic in one language, the appropriate claim is that the mechanism is real and severe where it occurs. We are not claiming to have estimated how often it occurs across the space of possible writing. We return to this scope question explicitly in Section~\ref{sec:limitations}.

\section{Related Work}
\label{sec:related}

Each element of this study intersects an established literature; the contribution is the specific intersection, and the counter-stereotypical, recorded-ground-truth design that connects them.

\subsection{Demographic inference from text}

That language models infer author attributes from text at scale is well established \citep{beyondmemorization,daiq,culturalsignal}. Recent audits document systematic demographic disparities in related tasks. \citet{berman2025attribution} introduce a fame- and demographically-balanced quote-attribution benchmark. They find large, systematic accuracy disparities by race, gender, and intersectional group across eleven models. The gap concentrates specifically in indirect, evidence-conditioned attribution rather than direct recall. \citet{islam2026message} shows LLMs generate systematically different persuasive framing depending on the demographic target of a message, indicating the bias runs in the generation direction as well as the inference direction. Closest in design logic to the present study, \citet{wyss2025unfair} audit authorship-attribution fairness on social-media text. They find no measurable demographic bias in a closed-world setting, but a clear tendency toward demographic-consistent misattribution once the true author is removed and the classifier must choose among incorrect candidates. This forced-choice-versus-open-set contrast closely parallels this paper's own matching-versus-open-ended contrast (Section~\ref{sec:f3}), obtained independently and on a different task.

What these studies share is scraped or synthetic text with self-reported or model-inferred demographic labels. The present study instead uses three items whose ages were recorded at collection. These items are counter-stereotypical by construction: the register-age relationship the scraped literature cannot isolate, because in unfiltered natural data, formal register and older authorship are often genuinely correlated.

\subsection{Register, not demography}

\citet{grieve2023register} argues, from parallel stylometric and multidimensional register analyses of human-authored text, that authorship signal is better explained by register variation than by dialect variation. Writers differ because they write in subtly different registers, not because they belong to different demographic speech communities. \citet{mikros2024crosslinguistic} show that a stylometric signal can survive machine translation across languages, suggesting the signal is at least partly abstracted above surface lexical choice. Read together, this literature supports treating register as a genuine, learnable authorial signature. The present study asks a different, behavioral question: do current LLMs treat register as a marker of \emph{identity} (who wrote this), or correctly localize it as a marker of \emph{register itself} (how formally this was written)? Findings~F1--F2 (Section~\ref{sec:results}) indicate that several of the six models conflate the two, and that three of them do so to the point of performing at or below chance.

\subsection{Confidence, calibration, and overconfidence}

LLM overconfidence and RLHF-induced miscalibration are documented findings \citep{chhikara2025confidence}. \citet{xie2026know} provide a mechanistic account: models trained by maximum-likelihood estimation (supervised fine-tuning) tend toward calibrated confidence, while reward-optimization methods (PPO, GRPO, DPO) induce overconfidence through reward exploitation instead. This paper's contribution is a trait-versus-signal decomposition. Rather than asking only whether confidence is miscalibrated in aggregate, we test, per model, whether stated confidence carries any within-model signal about correctness at all. For five of six models it does not, or it runs in the wrong direction. Between-model differences in mean confidence, meanwhile, are large and highly reliable. This reframes a common practice, reading a model's stated confidence as evidence about the answer, as uninformative about the answer for several current models, and informative mainly about which model was asked.

\subsection{Self-correction and its limits}

A growing literature shows LLM self-correction is limited without external feedback. \citet{li2025decomposing} decomposes self-correction into detection, localization, and correction sub-capabilities, and finds a counterintuitive Accuracy-Correction Paradox: weaker models intrinsically correct their own errors at a higher rate than stronger ones, whose errors run deeper and resist correction. \citet{kusner2026agentic} extend overconfidence findings to multi-step agentic settings, and related work \citep{selfcorrectionillusion,tacl2024survey} surveys the broader unreliability of intrinsic, feedback-free self-correction. These studies measure correction \emph{within} a single interaction. This paper measures a step largely absent from that literature: the content of a model's own retrospective explanation for an error, compared against that same model's accuracy and confidence in its other, independent sessions. Section~\ref{sec:f7} shows these can dissociate.

\subsection{Forced-choice evaluation}

Critiques of multiple-choice evaluation formats for LLMs argue that forced-choice accuracy can substantially overstate a model's actual competence relative to open-ended generation \citep{mcqaeval}. This paper demonstrates the effect directly, on identical stimuli, within a single study (Section~\ref{sec:f3}), rather than by comparing performance across different multiple-choice and open-ended datasets. This removes a common confound (item difficulty differing by format) from the comparison.

\section{Method}
\label{sec:method}

\subsection{Stimuli}

Three handwritten responses to the same free-response prompt (\emph{``What is happiness? Are you happy?''}, a community-college biology-course extra-credit item) served as the fixed stimulus set throughout. Author ages and genders were recorded at collection; the texts themselves are de-identified in all released materials. Each item is reproduced below in full, exactly as the writer submitted it, so readers can judge the register-age mismatch directly rather than from a paraphrase:

\begin{description}[nosep,leftmargin=1.2em]
\item[\texttt{concept} (ground truth: 17M).]
\begin{quote}\small Happiness is a concept that has numerous interpretations depending on the person. For example, some see happiness as being surrounded by people you care about, but generally, it is described as an emotion or state of being. While I am content with life and everything that comes with it, I am unsure if I am happy.\end{quote}
\item[\texttt{within} (ground truth: 28F).]
\begin{quote}\small I believe happiness comes from within yourself. It feels warm and as if you're `at home', more or less comfortable. yes, I would say I'm happy. I have a roof over my head a full belly, clothing, a car for transportation, a phone, etc.\end{quote}
\item[\texttt{hockey} (ground truth: 45M).]
\begin{quote}\small To me, happiness is having no worries. And season tickets to the Vegas Golden Knights hockey games, whom are in the NHL Playoffs this year. Yes, I am, even though I don't have season tickets, yet. Maybe.\end{quote}
\end{description}

All three are reproduced verbatim, including the writers' own grammar, capitalization, and punctuation; no text has been corrected or smoothed. The set is counter-stereotypical by design. The most formally hedged, abstract answer (\texttt{concept}) belongs to the youngest writer. The most casual, concrete, hobby-driven answer (\texttt{hockey}) belongs to the oldest.

\subsection{Models}

Six systems were tested via API, using the model identifiers under which each was accessed at the time of testing: Claude Fable~5, Gemini~3.1~Pro, GPT-5.6~Sol, Grok~4.5, DeepSeek~V4, and GPT-4o~(2024). The first five represent frontier or near-frontier systems as of mid-2026. GPT-4o~(2024) is included specifically as an earlier-generation anchor for the generational comparison in Finding~F8, not as a claim that it remains frontier. Exact API model strings and access dates are recorded in the released repository.

\subsection{Experimental conditions}

In every condition, a model was shown all three answers in one turn. It was asked to assign each to one of the three candidate labels (\texttt{17M}, \texttt{28F}, \texttt{45M}), along with a self-reported confidence (0--100) and a brief stated cue for each assignment. Assignments were required to be a one-to-one matching across the three items within a session. Six conditions varied what surrounded that request:

\begin{description}[nosep]
\item[A: Blind.] The three answers are presented with no source framing.
\item[B: Institutional label.] The answers are introduced as coming from ``a community-college biology final.''
\item[C: Open-ended.] No candidate labels are disclosed; the model free-generates an estimated age (a point value or a range) for each answer. Our analysis scores the 17-year-old's item. It is scored as correct if the true age (17) falls within a stated range, or within $\pm 3$ years of a stated point estimate.
\item[D: Audience-awareness instruction.] As in A, with one additional sentence instructing the model to consider that writers may adapt their register to their audience (e.g., a student writing for a teacher).
\item[E: Neutral label.] The answers are introduced with a topic-neutral framing (``a written survey''), holding institutional information constant at zero while still disclosing that the text was elicited for some purpose.
\item[P: Self-reported distribution.] As in A, but the model reports a percentage likelihood across all three candidate labels for each answer (e.g., \texttt{17M=48\% | 28F=6\% | 45M=46\%}), rather than a single top choice.
\end{description}

Every matching session in conditions A, B, and D (603 sessions) was immediately followed, within the same session, by a reveal turn. The true labels were disclosed, and the model was asked to explain what led it to its original assignments, correct or incorrect. This reveal-turn text is the basis for the post-reveal analysis in Finding~F7. Condition E sessions (99) have no recorded reveal text and are not part of that analysis.

Label order, the order in which the three answers were presented, was permuted across six orderings per model/condition cell to control for position effects, and each ordering is recorded in the \texttt{label\_order} column of the released data. Data were collected in seven batches between July 12 and July 14, 2026, each recorded in the \texttt{source\_file} column. Not all models appear in every batch. Gemini~3.1~Pro appears in only three of the seven batches and lost some sessions to API timeouts. Claude Fable~5 is absent from one batch and contributes only 39 rows to another. Effective $n$ is therefore smaller for these two models, most of all for Gemini. Per-model, per-batch row counts are reproduced by the released statistics script.

\subsection{Scale and data quality}

The full grid produced 1{,}081 API sessions and 3{,}243 item-level scored rows (three items per session). During post-hoc review, 18 rows (six sessions) were flagged for a known LLM API failure mode, duplicated model responses within a session, and excluded. All results below use the resulting analysis set of $N = 3{,}225$ rows from 1{,}075 sessions. The exclusion rule and every statistic in this paper, including all confidence intervals and significance tests, are reproduced by the released script \texttt{analysis/stats\_paper.py} run against the released master data file.

\subsection{Metrics}

For a set of scored trials, we report:

\begin{itemize}[nosep]
\item \textbf{Accuracy}: proportion of correct label assignments; chance under uniform random guessing among three labels is $1/3$.
\item \textbf{Misattribution@1} (for the \texttt{concept} item specifically): $1 - \text{accuracy}$ restricted to that item, i.e., the rate at which the 17-year-old's answer is assigned to the wrong age/gender bucket.
\item \textbf{Confidence--Correctness Gap (CCG)}: mean stated confidence (rescaled to $[0,1]$) minus accuracy, over the same set of trials. A positive CCG indicates systematic overconfidence.
\item \textbf{Session score}: the number of correct assignments in a matching session. Because assignment is one-to-one, a session can score only 0, 1, or 3. Under uniform random matching these occur with probability 2/6, 3/6, and 1/6 (mean 1, variance 1). Item rows within a session are therefore not independent, and all tests against chance are run on session scores.
\item \textbf{$r(\text{confidence}, \text{correct})$}: the Pearson point-biserial correlation between a model's stated confidence and a binary correctness indicator, computed within model and pooled across the four matching conditions (A, B, D, E). We additionally report the two-sided $p$-value for each correlation.
\item \textbf{Predictive entropy and $p(\text{truth})$} (Condition P only): Shannon entropy, in bits, of the self-reported three-way probability distribution (maximum $\log_2 3 \approx 1.585$ bits under a uniform distribution), and the probability mass placed on the correct label.
\item \textbf{Post-reveal text} (conditions A, B, D): coded once per session with pre-specified lexical patterns. The \emph{external-blame} flag marks language attributing the error to the environment or the text itself (e.g., phrasing to the effect that the text was misleading, or that the model stands by its original reasoning). A second count marks explanations containing the stem \emph{stereotyp-}. We treat both as lexically operationalized rather than independently multi-rater validated (Section~\ref{sec:limitations}).
\end{itemize}

All accuracy estimates are reported with Wilson score 95\% confidence intervals. Tests against chance use session scores: a $z$-test of the mean session score against 1, and a $\chi^2$ goodness-of-fit test of the 0/1/3 distribution against its chance distribution. Comparisons between conditions use Fisher's exact test, which remains valid when a cell is at 0 or 1. Item-level binomial $p$-values are not used for inference here, because they treat clustered rows as independent and are anticonservative.

\section{Results}
\label{sec:results}

\subsection{F1: Misattribution is current, severe, and in several cases at or below chance}

Table~\ref{tab:conda} reports Condition~A (blind presentation, pooled across the three items) for all six models. Claude Fable~5 and Gemini~3.1~Pro perform well above chance. GPT-5.6~Sol is above chance but well short of ceiling. The remaining three models average 0.69 to 1.05 correct per session against a chance mean of 1, while reporting 73--88\% mean confidence. The six models differ sharply in accuracy ($\chi^2(5) = 186.6$, $p = 2.0 \times 10^{-38}$ on item rows; Kruskal--Wallis on session scores $H = 84.0$, $p = 1.2 \times 10^{-16}$).

\begin{table}[h]
\centering
\scriptsize
\setlength{\tabcolsep}{2.5pt}
\caption{Condition A (blind). Accuracy is pooled across items with a Wilson 95\% CI. CCG = mean confidence $-$ accuracy. Misattr.@1 is the error rate on the 17-year-old's item. Sessions are split by score (0/1/3 correct). Mean is the mean session score (chance mean 1), tested against 1 by $z$-test ($p_{\mathrm{mean}}$).}
\label{tab:conda}
\begin{tabular}{lrcccccrcrc}
\toprule
Model & $n$ & Acc. & 95\% CI & Conf. & CCG & Misattr.@1 & Sess. & 0/1/3 & Mean & $p_{\mathrm{mean}}$ \\
\midrule
Claude Fable 5 & 108 & 0.94 & [0.88, 0.97] & 0.67 & $-$0.27 & 0.08 & 36 & 0/3/33 & 2.83 & $<$0.001 \\
Gemini 3.1 Pro &  51 & 0.76 & [0.63, 0.86] & 0.87 & +0.10 & 0.29 & 17 & 2/3/12 & 2.29 & $<$0.001 \\
GPT-5.6 Sol    & 120 & 0.63 & [0.54, 0.71] & 0.64 & +0.01 & 0.50 & 40 & 4/16/20 & 1.90 & $<$0.001 \\
Grok 4.5       & 126 & 0.35 & [0.27, 0.44] & 0.73 & +0.38 & 0.81 & 42 & 14/20/8 & 1.05 & 0.758 \\
GPT-4o (2024)  & 126 & 0.25 & [0.19, 0.34] & 0.86 & +0.60 & 0.81 & 42 & 24/11/7 & 0.76 & 0.123 \\
DeepSeek V4    & 126 & 0.23 & [0.17, 0.31] & 0.88 & +0.65 & 0.90 & 42 & 19/20/3 & 0.69 & 0.045 \\
\bottomrule
\end{tabular}
\end{table}

Because one-to-one matching makes the three item rows of a session dependent, we test against chance at the session level. Grok~4.5 is indistinguishable from chance on both the mean ($p = 0.76$) and the shape of its session distribution ($\chi^2$ goodness of fit, $p = 0.91$). This should be read as performance equivalent to guessing on this task, not as evidence that the model guesses uniformly at random. DeepSeek~V4's mean session score (0.69) is below chance ($p = 0.045$), though its distribution is not detectably different from the chance distribution ($p = 0.13$); we treat its below-chance performance as suggestive rather than established. GPT-4o~(2024) shows a different profile. Its mean is not detectably below chance ($p = 0.12$), but its session distribution departs from chance ($p = 0.003$): 24 of 42 sessions were entirely wrong, against 14 expected under random matching. For both models the error pattern is directional rather than noise, as F2 shows.

\subsection{F2: Same text, opposite readings}

Restricting to the \texttt{concept} item (the 17-year-old's answer) in Condition A only, the six models diverge sharply in how they resolve the same text (Table~\ref{tab:f2}). Claude Fable~5 assigns the correct label in 33 of 36 trials. The three lowest-accuracy models predominantly assign it to \texttt{28F}, with a substantial minority to \texttt{45M} and few to the correct \texttt{17M}. The pattern is legible as a folk theory rather than noise: formal, hedged phrasing reads as older. Stated cues in the raw transcripts are consistent with this: models attribute the teen's answer to considerations such as emotional maturity or life-stage framing that, empirically, ran in the wrong direction for this specific writer.

\begin{table}[h]
\centering
\small
\caption{Condition A only, \texttt{concept} item (17-year-old's answer): label assigned by each model, raw counts.}
\label{tab:f2}
\begin{tabular}{lrrr}
\toprule
Model & 17M (correct) & 28F & 45M \\
\midrule
Claude Fable 5 & 33 & 3  & 0  \\
Gemini 3.1 Pro & 12 & 2  & 3  \\
GPT-5.6 Sol    & 20 & 16 & 4  \\
Grok 4.5       & 8  & 22 & 12 \\
GPT-4o (2024)  & 8  & 25 & 9  \\
DeepSeek V4    & 4  & 25 & 13 \\
\bottomrule
\end{tabular}
\end{table}

\subsection{F3: Open-ended attribution is far worse than matching}
\label{sec:f3}

Removing the multiple-choice scaffold changes the picture substantially. Under Condition C, scoring an open-ended age estimate as correct if it falls within a stated range or within $\pm 3$ years of the true age (17), accuracy on the teen's item collapses for every model. Grok~4.5, GPT-4o~(2024), and DeepSeek~V4 score exactly $0.00$ ($n=42$ each). GPT-5.6~Sol reaches only $0.17$ ($n=40$). Even Claude Fable~5, which was near ceiling under matching, drops to $0.26$ ($n=31$). Gemini~3.1~Pro is highest at $0.33$ ($n=18$). That even the best matching-condition performer falls this far under open-ended elicitation indicates the matching-task accuracy in Table~\ref{tab:conda} does not reflect a stable, transferable estimate of the writer's age. It reflects, in part, an ability to rule out two disclosed alternatives, a substantially easier task than free estimation. This is consistent with prior findings that multiple-choice formats can mask the severity of a model's actual error rate \citep{mcqaeval}, here demonstrated on identical stimuli within a single study rather than across datasets that might differ in difficulty for unrelated reasons.

\subsection{F4/F5: Context inversion and a capability-gated fix}

Table~\ref{tab:context} reports accuracy under institutional framing (B), the audience-awareness instruction (D), and a neutral label (E), relative to the blind baseline (A).

\begin{table}[h]
\centering
\small
\caption{Accuracy under institutional framing (B), the audience-awareness instruction (D), and a neutral label (E), relative to blind (A). $p$-values are two-sided Fisher's exact tests against Condition A for the same model. Cell sizes (A/B/D/E): Claude 108/108/72/36; Gemini 51/51/54/45; GPT-5.6 120/63/87/39; Grok 126/126/87/54; GPT-4o 126/126/90/54; DeepSeek 126/126/90/54. GPT-5.6 Sol's Condition B $n$ is reduced by abstentions (Section~\ref{sec:f8}).}
\label{tab:context}
\begin{tabular}{lcccccccc}
\toprule
Model & Acc.\ A & Acc.\ B & $p$(B) & Acc.\ D & $p$(D) & Acc.\ E & $p$(E) \\
\midrule
Claude Fable 5 & 0.94 & 0.69 & $<$0.001 & 1.00 & 0.082 & 0.94 & 1.000 \\
Gemini 3.1 Pro & 0.76 & 0.61 & 0.135 & 0.89 & 0.121 & 0.64 & 0.261 \\
GPT-5.6 Sol    & 0.63 & 0.54 & 0.266 & 0.84 & 0.002 & 0.67 & 0.848 \\
Grok 4.5       & 0.35 & 0.33 & 0.894 & 0.36 & 1.000 & 0.31 & 0.732 \\
GPT-4o (2024)  & 0.25 & 0.29 & 0.670 & 0.22 & 0.631 & 0.46 & 0.008 \\
DeepSeek V4    & 0.23 & 0.21 & 0.761 & 0.18 & 0.398 & 0.30 & 0.354 \\
\bottomrule
\end{tabular}
\end{table}

Two effects reach conventional significance. Institutional framing significantly \emph{degrades} the strongest model (Claude Fable~5: $0.94 \rightarrow 0.69$, $p < 0.001$). The audience-awareness instruction significantly \emph{improves} GPT-5.6~Sol by 21 points ($0.63 \rightarrow 0.84$, $p = 0.002$). It also moves Claude Fable~5 to ceiling (72 of 72), but with the baseline already at 0.94 this change is not statistically detectable at this sample size ($p = 0.082$). For the three lowest-accuracy models (Grok, GPT-4o, DeepSeek), neither manipulation produces a detectable change. We read this as a floor effect rather than an absence of the phenomenon: these models are already at or near chance, leaving little room for a context manipulation to move performance in a way this sample size can detect.

The direction of the effect that \emph{is} detectable is nonetheless notable. Institutional metadata (``community-college biology final'') significantly damages the one model otherwise near ceiling, while a topic-neutral label carries no such cost (Condition E accuracy for Claude Fable~5 is indistinguishable from Condition A). This is consistent with context intended to describe a setting being read as a claim about the people in it, a low-stakes instance of the mechanism behind socioeconomic or institutional stereotyping. We did not test this mechanism directly.

One result runs the other way. For GPT-4o~(2024), the neutral survey label raised accuracy from 0.25 to 0.46 ($p = 0.008$), the only significant Condition E effect in the study. Its Condition E cell is small ($n = 54$) and the effect does not appear for any other model, so we report it without interpretation, as a candidate for replication.

\subsection{F6: Confidence is a per-model trait, not a per-item signal}

\begin{table}[h]
\centering
\small
\caption{Confidence--correctness correlation, pooled across matching conditions A/B/D/E, with mean and SD of stated confidence (0--100) to indicate range restriction.}
\label{tab:conf}
\begin{tabular}{lrrrrr}
\toprule
Model & $n$ & Conf. mean & Conf. SD & $r$(conf., correct) & $p$ \\
\midrule
Claude Fable 5 & 324 & 67.7 &  7.0 & +0.119 & 0.032 \\
Gemini 3.1 Pro & 201 & 88.3 &  5.7 & $-$0.143 & 0.043 \\
GPT-5.6 Sol    & 309 & 62.4 & 12.6 & $-$0.015 & 0.790 (n.s.) \\
Grok 4.5       & 393 & 73.3 &  8.0 & $-$0.207 & $<$0.001 \\
GPT-4o (2024)  & 396 & 84.6 &  7.6 & $-$0.273 & $<$0.001 \\
DeepSeek V4    & 396 & 88.0 &  5.2 & +0.015 & 0.762 (n.s.) \\
\bottomrule
\end{tabular}
\end{table}

Table~\ref{tab:conf} shows the within-model correlation between stated confidence and correctness, pooled across the four matching conditions. The correlation is significant and \emph{positive} only for Claude Fable~5 ($r=+0.12$, $p=0.032$), weak, but in the calibrated direction. It is significant and \emph{negative} for Gemini~3.1~Pro, Grok~4.5, and GPT-4o~(2024) (all $p<0.05$, the latter two $p<0.001$): these models report \emph{higher} confidence specifically when wrong. For GPT-5.6~Sol and DeepSeek~V4 the correlation is statistically indistinguishable from zero.

This null result should not be read as evidence of calibration. DeepSeek~V4 has the narrowest confidence range in the study (SD 5.2 points on a 0--100 scale, mean 88.0), so there is little within-model variation for a correctness signal to correlate with in the first place. A model whose confidence output is close to a constant cannot, by construction, produce a strong correctness correlation regardless of its actual reliability. Gemini~3.1~Pro has a similarly narrow range (SD 5.7) and still shows a significant negative correlation, so range restriction limits, but does not by itself remove, a detectable signal. The appropriate reading for DeepSeek is that its stated confidence carries almost no per-item information, rather than that it is well calibrated. Between-model differences in mean confidence (62.4 to 88.3 across conditions A/B/D/E) are large by comparison. Across this dataset, a model's stated confidence predicts \emph{which model produced it} far more reliably than whether the underlying answer is correct.

\subsection{F7: Confession without correction}
\label{sec:f7}

Every matching session in conditions A, B, and D ended with a reveal turn in which ground truth was disclosed and the model was asked to explain its reasoning. Each session produces one explanation, so we code these per session (603 sessions). Language attributing the error externally, to the text or the setup, is rare: 4 of 603 sessions, 3 of them DeepSeek~V4's (Table~\ref{tab:blame}; Fisher's exact test, DeepSeek against the pooled remaining five models: 3/114 vs.\ 1/489, odds ratio $= 13.2$, $p = 0.023$).

\begin{table}[h]
\centering
\small
\caption{Post-reveal explanations (conditions A, B, D), coded per session. ``Ext.\ blame'' counts sessions matching the external-blame pattern; ``Stereotyping'' counts explanations containing the stem \emph{stereotyp-}; ``Thanks'' counts explanations that open with \emph{thank}.}
\label{tab:blame}
\begin{tabular}{lrrrr}
\toprule
Model & Sessions & Ext.\ blame & Stereotyping & Thanks \\
\midrule
Claude Fable 5 &  96 & 0 & 18 &   0 \\
Gemini 3.1 Pro &  52 & 0 & 10 &   0 \\
GPT-5.6 Sol    & 114 & 1 & 16 &   0 \\
Grok 4.5       & 113 & 0 &  5 &   0 \\
GPT-4o (2024)  & 114 & 0 &  2 &   0 \\
DeepSeek V4    & 114 & 3 & 68 & 108 \\
\bottomrule
\end{tabular}
\end{table}

The more striking contrast is between what DeepSeek~V4 says and what it does. Its explanations are the study's most self-critical. In 68 of 114 the word stem \emph{stereotyp-} appears, far more than for any other model, and the passages we read use it about the model's own reasoning (for example, acknowledging that it had stereotyped sports enthusiasm and a grammatical slip as adolescent). The count is lexical and does not by itself separate self-blame from other uses of the word. In 108 of 114, the explanation opens by thanking the user. Yet in its independent sessions under blind presentation, DeepSeek~V4 shows the lowest accuracy and the highest stated confidence of any model (Table~\ref{tab:conda}).

Because each API session is stateless, no transfer from one session's explanation to another session was possible, and we do not claim the model failed to learn. The finding is narrower. The quality of a model's self-diagnosis does not track the quality of its judgment: the model that most fluently names the bias mechanism is the model most governed by it. Post-hoc self-explanation, on this evidence, is better read as a genre of plausible text generation than as a readout of the process that produces the model's judgments. This extends prior findings that intrinsic self-correction is limited within a single interaction \citep{li2025decomposing,kusner2026agentic,selfcorrectionillusion}. A within-session test (re-presenting a new ordering after the reveal, in the same context) would measure whether such diagnosis changes behavior when it can, and is a natural extension.

\subsection{F8: A generational comparison}
\label{sec:f8}

Comparing GPT-4o (2024) against GPT-5.6~Sol (2026), the same lab's model roughly two years apart, shows movement on every metric in this study. Accuracy on the matching task rose from 0.25 to 0.63 (Fisher's exact $p = 2.2 \times 10^{-9}$). Mean confidence fell from 0.86 to 0.64. The confidence--correctness gap fell from $+0.60$ to $+0.01$. Abstention, declining to assign labels rather than committing to an incorrect one, rose from none to 15 sessions (45 item rows), including responses stating outright that age or gender cannot reliably be inferred from an anonymous writing sample. Every abstention was a full refusal of all three labels, and 12 of the 15 occurred under institutional framing (Condition B), which reduces GPT-5.6~Sol's Condition B sample to 63 scored rows.

This pattern is consistent with a deliberate training-time shift toward calibration and appropriate abstention. It is not, however, a general trend across labs tested at a similar or later point in time: DeepSeek~V4, among the newer models in this study, has the lowest accuracy under blind presentation and confidence among the highest in every condition, and it never abstained. Generational progress on this failure mode appears to be lab-specific and trainable where a lab targets it, rather than an automatic consequence of model scale or recency.

\subsection{F9: A meta-finding on tool fallibility}

During the conduct of this study, AI-assisted tooling used for coding and analysis produced several errors that mirror the study's own subject matter: a fabricated code-review summary, a silently altered answer key, an invented quotation offered as supporting evidence, and, via a parser defect introduced by the author, one transient false result in an intermediate analysis pass. Each was caught by manual human verification before inclusion in the released dataset or this paper, and none affects the results reported above. We disclose this not as a caveat to be minimized but as directly relevant methodological information. A study of confident, fluent error in language models encountered confident, fluent error, from language models, at multiple points in its own production. We recommend that studies of this kind budget explicitly for adversarial human review of AI-assisted analysis steps, and report what that review caught.

\subsection{Condition P: self-reported distributions}

When asked for a probability over all three labels rather than a single choice, models broadly reproduce the ranking seen under forced choice (Table~\ref{tab:p}; only Grok~4.5 and GPT-4o swap places). The mass placed on the true label ranges from 25.0\% (DeepSeek~V4) to 61.4\% (Gemini~3.1~Pro). Mean entropy stays between 0.98 and 1.34 bits for every model (maximum 1.58), including models whose top choice is wrong in most trials. Expressing uncertainty as a distribution does not, on this task, make the weak models more accurate or the strong ones less so.

\begin{table}[h]
\centering
\small
\caption{Condition P. Top-1 accuracy (Wilson 95\% CI), mean probability on the true label, and mean predictive entropy (bits; maximum 1.58).}
\label{tab:p}
\begin{tabular}{lrcccc}
\toprule
Model & $n$ & Top-1 & 95\% CI & $p$(truth) & Entropy \\
\midrule
Claude Fable 5 & 68 & 0.99 & [0.92, 1.00] & 58.8\% & 1.34 \\
Gemini 3.1 Pro & 54 & 0.76 & [0.63, 0.85] & 61.4\% & 0.98 \\
GPT-5.6 Sol    & 87 & 0.62 & [0.52, 0.72] & 47.6\% & 1.25 \\
Grok 4.5       & 84 & 0.27 & [0.19, 0.38] & 30.6\% & 1.29 \\
GPT-4o (2024)  & 87 & 0.34 & [0.25, 0.45] & 34.3\% & 1.34 \\
DeepSeek V4    & 87 & 0.18 & [0.12, 0.28] & 25.0\% & 1.11 \\
\bottomrule
\end{tabular}
\end{table}

\section{Discussion}

The unifying pattern across F1--F7 is that several current frontier models treat linguistic register as a demographic signal, rather than correctly localizing it as a fact about how, not who, a passage was written. This is precisely backwards on the one item in this study engineered to make the two come apart, and the models that make this error most severely do so with reported confidence in the 73--88\% range, not with hedged uncertainty. For any deployed system that infers or personalizes based on writer demographics inferred from style (content moderation, ad or content personalization, sock-puppet and duplicate-account detection, resume or forum-post screening), this is the same failure mode operating silently on real users rather than on a disclosed research stimulus.

The confidence findings (F6) have a direct implication for how model output should be consumed downstream. For five of the six models tested, a per-item confidence score is either uncorrelated with correctness or runs the wrong way. Using it as though it were informative, for example routing low-stated-confidence answers to human review while trusting high-stated-confidence ones, would, for Gemini, GPT-4o, and Grok specifically, systematically misroute in the wrong direction, since confidence and correctness are \emph{negatively} associated for those models within this task.

Findings F4/F5 together suggest that prompting can activate calibration that already exists in a model, but cannot manufacture it from nothing. The audience-awareness sentence is a large, significant lever for GPT-5.6~Sol, moves Claude Fable~5 from 0.94 to ceiling (not statistically detectable at this sample size), and is a measurable non-event for the three weakest models. Read alongside \citet{xie2026know}'s finding that training objective, not prompting, is the primary determinant of calibration reliability, this points toward prompting as a tool for eliciting latent capability rather than a substitute for it, consistent with, but not identical to, that paper's training-time account.

Finding F7 is the study's sharpest caution for anyone using a model's own retrospective explanation as evidence of anything beyond the immediate turn. A model that produces extensive, specific self-diagnosis of its own error mechanism, and is nonetheless the least accurate and most confident model under blind presentation, shows that fluent self-explanation and quality of judgment are dissociable in current systems. Evaluation protocols that treat the quality of a model's self-correction language as a proxy for reliability should, on this evidence, be treated with skepticism until a causal link is separately demonstrated.

\section{Limitations}
\label{sec:limitations}

This section is intended to be read before the results above are cited elsewhere, not after.

\begin{itemize}[nosep]
\item \textbf{Scope.} Three stimulus items, one topic (a happiness/wellbeing free-response), one language (English). This is a structured case study demonstrating that a mechanism exists and can be severe; it is not a population-level estimate of how often language models misattribute author demographics from text in general, and the specific accuracy numbers in Table~\ref{tab:conda} should not be extrapolated as base rates for other topics, writing styles, or languages.
\item \textbf{Non-independence.} The three item rows of a session share a model, a run, and a label ordering, and one-to-one matching makes their correctness dependent. Tests against chance are therefore run on session scores, and post-reveal text is coded once per session. Between-condition comparisons and the confidence--correctness correlations still pool item rows and should be read with this clustering in mind; a mixed-effects model with a session random effect is a natural extension.
\item \textbf{Statistical power at the floor.} The Fisher tests in Table~\ref{tab:context} detect the large effects present (Claude Fable~5 under B, GPT-5.6~Sol under D), but are underpowered, at this sample size, to distinguish a true null from a real-but-small context effect for the three lowest-accuracy models. Absence of significance there is not evidence of absence of the phenomenon.
\item \textbf{Post-reveal coding.} The external-blame flag and the counts in Table~\ref{tab:blame} are pre-specified lexical pattern matches over the post-reveal text, reproduced by the released script. They are not an independently validated multi-rater qualitative code; inter-rater reliability was not computed. The \emph{stereotyp-} count in particular does not distinguish self-blame from other uses of the word.
\item \textbf{Reveal-turn coverage.} Post-reveal text is available for conditions A, B, and D only (603 of 702 matching sessions). Condition E sessions have no recorded reveal text.
\item \textbf{Uneven cell sizes.} Gemini~3.1~Pro appears in only three of the seven data batches and lost some sessions to API timeouts. Claude Fable~5 is absent from one batch and only partly present in another. Their $n$ is smaller than that of the other models, which widens their confidence intervals correspondingly (visible in Table~\ref{tab:conda}).
\item \textbf{Human baseline.} A human-rater baseline on the same matching task, which would contextualize how much of this failure is specifically about language models versus a harder-than-it-looks task for any judge, is not included here.
\item \textbf{API vs.\ product behavior.} All models were accessed directly via API with a fixed system framing; consumer chat products built on the same underlying models may apply additional behavior layers (safety fine-tuning, system prompts, retrieval) not present in this setup, and may not reproduce these exact numbers.
\item \textbf{Analysis tooling.} As disclosed in F9, AI-assisted tools were used during data processing and produced several errors during the study, each caught by manual verification. We consider this a transparency disclosure rather than a defect specific to this dataset, but it is a relevant caveat for readers evaluating the provenance of any single number above; the released raw outputs and analysis script allow independent re-verification of every value in this paper.
\end{itemize}

\section{Ethics Statement}

The three stimulus texts are handwritten extra-credit answers to a community-college biology final, collected in 2025. Each writer recorded an age and gender at collection. The texts are de-identified in all released materials, and no personally identifying information beyond age and gender is retained or published. Age and gender categories used here (a specific two-digit age paired with a binary gender label per writer) reflect how ground truth was recorded at collection and are not intended as a general demographic taxonomy. The answers were written by hand as part of a class activity, and the writers signed them. The study idea arose after collection. The signed originals were later lost and cannot be produced, and no research-specific consent documentation or formal ethics review exists. Retrospective written consent for research use and publication is being sought from the writers; any text whose writer declines will be removed from the released data and from subsequent versions of this paper. This study evaluates model behavior on three short existing texts and involves no further interaction with the writers. All data, code, and raw model outputs are released under CC~BY~4.0 (data/text) and MIT (code) to support independent verification.

\section{Conclusion}

Six frontier language models were asked to determine which of three real people, whose ages and genders were recorded, wrote which of three real, short pieces of text. On the one item constructed so that formal register and youth coincide, most models got it wrong and said so confidently. When shown their error and asked to explain it, at least one model produced a detailed, fluent account of exactly what went wrong, while showing the lowest accuracy and highest confidence of any model under blind presentation.

Each of these results is drawn from a small, deliberately constructed case study rather than a large natural sample, and should be read at that scope. Within that scope, the findings are clear: for several current systems, linguistic register functions as an unacknowledged demographic proxy, stated confidence is not a reliable per-item signal of correctness, and the quality of post-hoc self-explanation does not track the quality of judgment. All three are directly relevant to any deployed system that infers, personalizes, or moderates based on the demographic characteristics of a piece of writing.

\section*{Data and Code Availability}
Data, analysis code, and raw model outputs: \url{https://github.com/BraveAnn011/llm-author-misattribution}. Data and text are released under CC~BY~4.0; code is released under the MIT license.

\end{document}